\documentclass[conference]{ieeeconf}
\IEEEoverridecommandlockouts
\usepackage{cite}
\usepackage{amsmath,amssymb,amsfonts}
\usepackage{algorithmic}
\usepackage{graphicx}
\usepackage{textcomp}
\usepackage{xcolor}
\usepackage{subcaption}
\usepackage{multirow}
\usepackage{soul}
\usepackage{etoolbox}

\usepackage[algo2e,algoruled,boxed,lined,norelsize, vlined, linesnumbered]{algorithm2e}
\let\oldnl\nl
\newcommand{\nonl}{\renewcommand{\nl}{\let\nl\oldnl}}

\def\BibTeX{{\rm B\kern-.05em{\sc i\kern-.025em b}\kern-.08em
    T\kern-.1667em\lower.7ex\hbox{E}\kern-.125emX}}
\usepackage[bookmarks=true,breaklinks]{hyperref}

\newtoggle{editmode}
\newcommand{\Add}[1]{\iftoggle{editmode}{\textcolor{red}{#1}}{#1}}
\newcommand{\Cross}[1]{\iftoggle{editmode}{\st{#1}}{\ignorespaces}}

\title{\LARGE \bf Enhancing Task Performance of Learned Simplified Models via Reinforcement Learning
\thanks{\Add{$^\ast$Toyota Research Institute provided funds to support this work.}}
\thanks{$^{1}$The authors are with the GRASP Laboratory, University of Pennsylvania,
	Philadelphia, PA 19104, USA \{xuanhien, posa\}@seas.upenn.edu}
}

\author{Hien Bui$^{1}$ and Michael Posa$^{1}$}
\begin{document}
\maketitle

\begin{abstract}
In contact-rich tasks, the hybrid, multi-modal nature of contact dynamics poses great challenges in model representation, planning, and control.
Recent efforts have attempted to address these challenges via data-driven methods, learning dynamical models in combination with model predictive control.
Those methods, while effective, rely solely on minimizing forward prediction errors to hope for better task performance with MPC controllers. This weak correlation can result in data inefficiency as well as limitations to overall performance.
In response, we propose a novel strategy: using a policy gradient algorithm to find a simplified dynamics model that explicitly maximizes task performance.
Specifically, we parameterize the stochastic policy as the perturbed output of the MPC controller, thus, the learned model representation can directly associate with the policy or task performance. 
We apply the proposed method to contact-rich tasks where a three-fingered robotic hand manipulates previously unknown objects. Our method significantly enhances task success rate by up to 15\% in manipulating diverse objects compared to the existing method while sustaining data efficiency. Our method can solve some tasks with success rates of 70\% or higher using under 30 minutes of data.
All videos and codes are available at \href{https://sites.google.com/view/lcs-rl}{https://sites.google.com/view/lcs-rl}.
\end{abstract}

\section{Introduction}
 \begin{figure}[t]
    \centering
    \includegraphics[width=0.7\linewidth]{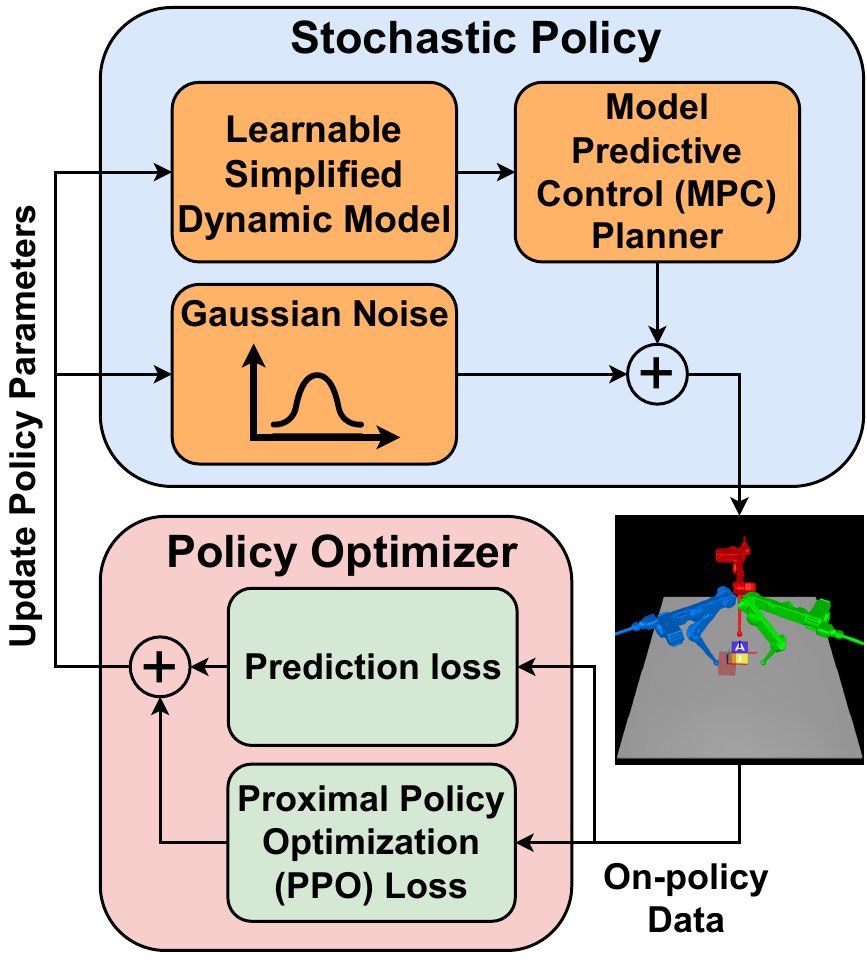}
    \caption{The diagram demonstrates our proposed framework of learning simplified dynamic models for solving contact-rich manipulation tasks in low data regimes. Our framework proposes an iterative learning loop that consists of main components: a stochastic policy and a policy optimizer. \textbf{Top panel}: Using learned dynamic models under MPC scheme and Gaussian noise to construct the stochastic policy. \textbf{Bottom panel:} Combining PPO and prediction loss to optimize the policy parameters with the collected on-policy data.}
	\label{img.overall_framework}
 \vspace{-15pt}
\end{figure}
In many robotics tasks, such as dexterous manipulation and locomotion, robots frequently need to make and break contact with the environment.
Yet, finding explicit models and policies that can exploit the hybrid complex interaction of the robot with its environment to solve the tasks remains a challenge.
Some works in model-based control \cite{Marcucci.Tedrake2019, Frick.etal2019, Aydinoglu.etal2023, Aydinoglu.Posa2022b} have attempted to explicitly identify contact modes and plan the contact sequences.
However, these approaches face scalability challenges as the number of contact modes increases.

Recent data-driven methods have made significant advances in tackling that scalability issue, broadly offering two primary strategies.
Modern model-free reinforcement learning (RL) directly parameterizes control policies with deep neural networks, then iteratively improves the policies through large-scale trial and error \cite{Schulman.etal2017, Fujimoto.etal2018, Haarnoja.etal2018}.
However, because of their data inefficiency, carrying out experiments on real robotic systems is always resource-intensive and time-consuming.
In contrast, \Cross{a line of work, often referred to as model-based RL} \Add{a large portion of model-based RL}\cite{Nagabandi.etal2018, Chua.etal2018, Nagabandi.etal2020, Morgan.etal2021, Ebert.etal2018, Zhang.etal2019, Ghugare.etal2022} leverages the expressiveness power of deep neural networks to learn intricate dynamic models. 
The learned models are subsequently employed for trajectory planning\Add{/predictive control, typically} via random shooting techniques.
Despite improving data efficiency compared to model-free RL, model-based RL remains data-intensive because conventional methods for learning deep dynamic models struggle to capture stiffness and multi-modal contact dynamics \cite{Parmar.etal2021a, Bianchini.etal2022}.
Moreover, by adopting simpler model representations such as time-varying linear or Gaussian Process models, some works \cite{Kumar.etal2016, Levine.Koltun2013, Deisenroth.Rasmussen2011} demonstrate good performance with limited data.

The most recent work \cite{Jin2024} shows great performance with only a few minutes of data by learning \Add{a linear complementarity system (LCS), a piecewise-affine and reduced-order representation of multi-contact dynamics}, and combining it with an MPC planner. However, two main building blocks of this work, the dynamic model fitting and planning with the learned model, appear as two de-coupled optimization problems repeated over many cycles of learning. More specifically, ensuring better forward prediction capability of the learned dynamic model is necessary but might not be sufficient to achieve better task performance, leading to limitations of data efficiency and task performance. This issue is known as \textit{objective mismatch} \cite{Lambert.etal2021}.

This paper presents LCS-RL, a low-dimensional RL framework, to address the above issue, aiming to further enhance task performance and data efficiency. Particularly, the proposed framework directly bridges the dynamic model learning part to task performance optimization via a policy gradient algorithm.

\subsection{Contributions}
\begin{enumerate}
    \item We present LCS-RL, a novel framework that leverages the combination of RL and simple multi-contact models for solving contact-rich tasks. Specifically, our framework applies a reinforcement learning algorithm to directly maximize the task performance of simplified models in combination with the MPC planner.
   \item We show that the proposed method consistently achieves higher task performance, up to $15\%$ in three-fingered robot manipulation tasks with various objects compared to the prior methods \cite{Jin2024, Nagabandi.etal2020}. In addition, our method is data efficient as it can solve some dexterous manipulation tasks with $70\%$ to $96\%$ success rates using just under $30$ minutes of data.
   \item We also demonstrate that the learned LCS model of one object can be transferred to other objects, drastically improving data efficiency.
\end{enumerate}

\section{Related Work}
\subsection{Differentiable MPC and Reinforcement Learning}
\Add{Control policies formulated as differentiable MPC problems and optimized using backpropagation, either through imitation learning or RL loss, have been extensively studied \cite{Okada.etal2017, Amos.etal2019, Esfahani.etal2021, Saxena.etal2021, Jin_NEURIPS.etal2020, Jin_NEURIPS.etal2021, wan2024difftop, pineda2022theseus, xu2023revisiting}.}
\Add{Amos et al. \cite{Amos.etal2019}, Xu et al. \cite{xu2023revisiting}, and Jin et al. \cite{Jin_NEURIPS.etal2020, Jin_NEURIPS.etal2021} propose using either a differential set of Karush–Kuhn–Tucker (KKT) or Pontryagin’s maximum principle (PMP) conditions to ensure MPC problem differentiability, with limited experiments in low-dimensional tasks.}
\Cross{The prior works\mbox{\cite{Okada.etal2017, Amos.etal2019, Esfahani.etal2021}} are the most relevant to ours: forming parametric MPC with learnable either cost functions or dynamic constraints, and optimizing its parameters via some losses such as imitation or RL loss.}
\Add{Recent works \cite{Esfahani.etal2021, Saxena.etal2021, wan2024difftop} extend these methods to more complex robotics problems. Esfahani et al. \cite{Esfahani.etal2021} use a specialized Q-learning algorithm to learn MPC cost function parameters, effective in mobile robot tasks but critically requiring ground-truth dynamics. Wan et al. \cite{wan2024difftop} focus on image-based tasks, introducing a differentiable sampling-based MPC policy to learn latent dynamics models, encoder, and Q-value predictors concurrently, but requiring substantial data.}
\Add{For data-efficient system identification and control in state-based contact-rich tasks, Saxena et al. \cite{Saxena.etal2021} are the most relevant. They parameterize the dynamics model using a switching linear dynamical model with a contact-mode prediction function and construct a differentiable feedback controller (LQR), optimizing both its cost matrices and dynamics model parameters to match expert demonstrations.}
\Cross{However, those works express some limitations. Amos et al. \mbox{\cite{Amos.etal2019}} propose to use simple linear dynamic models and limit their experiments to imitation learning on low-dimensional tasks.}
\Cross{Meanwhile, Esfahani et al. \mbox{\cite{Esfahani.etal2021}} focus more on robust MPC, optimizing parameters of the cost functions for better performance with the presence of disturbances and uncertainties, but critically require access to ground-truth dynamics.}

\subsection{\Add{Main} Baseline for Comparisons}
Jin et al. \cite{Jin2024} suggest that there exists a simple model that can adequately capture task-relevant contact dynamics, thereby enabling both high performance and real-time control for contact-rich manipulation.
In particular, the authors propose to use a reduced-order hybrid model to represent and use the model predictive controller for planning. Since the model is far simpler than deep neural networks, much less data is required for model learning. Their framework achieves high task performance with under 5 minutes of data.
In this paper, we compare the task performance and data efficiency of our proposed method against this baseline in some dexterous manipulation tasks.

\section{Backgrounds}

\subsection{Linear Complementarity Systems}
A discrete-time linear complementarity system (LCS) is a piecewise-affine system, where the state evolution is governed by linear dynamics in (\ref{eqn.lcs_1}) and a linear complementarity problem (LCP) in (\ref{eqn.lcs_2}).
\begin{subequations}
	\label{eqn.lcs}
	\begin{align}
	\boldsymbol{x}_{t+1}=A \boldsymbol{x}_t + B \boldsymbol{u}_t + C \boldsymbol{\lambda}_t + \boldsymbol{d},\label{eqn.lcs_1}\\
	0 \leq \boldsymbol{\lambda}_t \perp D \boldsymbol{x}_t + E \boldsymbol{u}_t + F \boldsymbol{\lambda}_t + \boldsymbol{c} \geq 0.
	\label{eqn.lcs_2}
	\end{align}
\end{subequations}
Here, $\boldsymbol{x}_t \in \mathbb{R}^{n_x}$, $\boldsymbol{u}_t \in \mathbb{R}^{n_u}$, and $\boldsymbol{\lambda}_t \in \mathbb{R}^{n_{\lambda}}$ are respectively the system state, action, and the complementarity variable at time step $t$. And, $\boldsymbol{x}_{t+1} \in \mathbb{R}^{n_x}$ is the system state at the next time step $t+1$.
\Add{The symbol $\perp$ denotes zero inner product or orthogonality of two vectors.}
Moreover, the matrix $A \in \mathbb{R}^{n_x \times n_x}$ defines the autonomous dynamics and matrix $B \in \mathbb{R}^{n_x \times n_u}$ captures the effect of actions on states. And, the matrix $C \in \mathbb{R}^{n_x \times n_{\lambda}}$ and $\boldsymbol{d} \in \mathbb{R}^{n_x}$ describe the effect of the contact forces and the constant forces acting on the state respectively. Other matrices $D \in \mathbb{R}^{n_{\lambda} \times n_x}, E \in \mathbb{R}^{n_{\lambda} \times n_u}, F \in \mathbb{R}^{n_{\lambda} \times n_{\lambda}}$ and $\boldsymbol{c} \in \mathbb{R}^{n_{\lambda}}$ altogether capture the relationship between states, actions, and contact forces.
\Add{The LCS models are commonly used in modeling multi-contact robotics problems \cite{Stewart.etal2000, Aydinoglu.etal2022}.}

\subsection{Learning Linear Complementarity Systems}
\label{sec.learning_lcs}
Given a data buffer $\mathcal{D}$ that contains some state transitions $(\boldsymbol{x}_t, \boldsymbol{u}_t, \boldsymbol{x}_{t+1})$, we can learn all matrix and vector parameters of an LCS model $(A, B, C, \boldsymbol{d}, D, E, F, \boldsymbol{c})$ in (\ref{eqn.lcs}) by using the gradient descent method with the violation-based loss, proposed by Jin et al. \cite{Jin.etal2022}
\begin{equation}
	\resizebox{0.9\columnwidth}{!}{$\displaystyle
	\begin{aligned}
	&\mathcal{L}_{\mathrm{vio}}^{\boldsymbol{\Theta}} =\min _{\boldsymbol{\lambda}_t \geq 0, \boldsymbol{\phi}_t \geq 0} \frac{1}{2}\left\|A \boldsymbol{x}_t + B \boldsymbol{u}_t + C \boldsymbol{\lambda}_t+\boldsymbol{d}-\boldsymbol{x}_{t+1}\right\|^2 \\
	& +\frac{1}{\xi}\left(\boldsymbol{\lambda}_t^{\mathrm{T}} \boldsymbol{\phi}_t+\frac{1}{2 \gamma}\left\|D \boldsymbol{x}_t + E \boldsymbol{u}_t + F \boldsymbol{\lambda}_t+\boldsymbol{c}-\boldsymbol{\phi}_t\right\|^2\right).
	\end{aligned}$}
	\label{eqn.vio_loss}
\end{equation}
Particularly, the loss $\mathcal{L}_{\mathrm{vio}}^{\boldsymbol{\Theta}}$ itself is an optimization problem whose first and second terms specify the violation of the affine dynamics (\ref{eqn.lcs_1}) and the LCP constraint (\ref{eqn.lcs_2}), respectively.
Under the condition $0 < \gamma \leq \sigma_{\text{min}}(F + F^T)$, finding $\boldsymbol{\lambda}_t$ to minimize the second term is equivalent to directly solving an LCP in (\ref{eqn.lcs_2}) for $\boldsymbol{\lambda}_t$, but poses a better-conditioned landscape for $\mathcal{L}_{\mathrm{vio}}^{\boldsymbol{\Theta}}$, thus enabling the identification of multi-modal and stiff dynamics.
The hyper-parameter $\xi > 0$ aims to balance two terms of the loss $\mathcal{L}_{\mathrm{vio}}^{\boldsymbol{\Theta}}$; and $\boldsymbol{\phi}_{t} \in \mathbb{R}^{n_{\lambda}}$ is an introduced slack variable for the complementarity equation.
Full explanations of the loss formulation and its hyper-parameters can be found in \cite{Jin.etal2022}.

As proven in \cite{Jin.etal2022}, using Envelope Theorem \cite{Afriat1971}, we can analytically compute the gradient of the violation-based loss with respect to LCS parameters $\frac{d \mathcal{L}^{\boldsymbol{\Theta}}_{\mathrm{vio}}}{d \boldsymbol{\Theta}}$ without differentiating through the
solution of the optimization problem.
\begin{algorithm2e}[t!]
	\SetAlgoLined
    \nonl \textbf{Parameterization:} LCS-MPC policy parameters $\boldsymbol{\theta}$;\\
    \nonl \textbf{Hyper-parameters:} The number of warm-up \\ 
    \nonl iterations $M$; the number of policy improvement \\ 
    \nonl steps $N_p$; and learning rate $\eta$;\\
    \nonl \textbf{Initialization:} $\boldsymbol{\theta}_0$, empty data buffer $\mathcal{D}$;\\
	\For{$k=0, 1,..., M$}
	{
		{Collect $N$ rollout trajectories by running the LCS-MPC stochastic policy $\pi_{\boldsymbol{\theta}_{k}}$ and add to $\mathcal{D}$
		}
  
		\For{$i\gets0$ \KwTo $N_p$}{
			Using data in $\mathcal{D}$, compute the gradient $\frac{d \mathcal{L}_{\mathrm{vio}}^{\boldsymbol{\theta}_i}}{d \boldsymbol{\theta}_i}$ \\
			Update $\boldsymbol{\theta}_{i+1}\leftarrow \boldsymbol{\theta}_{i} - \eta \frac{d \mathcal{L}_{\mathrm{vio}}^{\boldsymbol{\theta}_i}}{d \boldsymbol{\theta}_i}$
		}
	}	
	Save the final parameters $\boldsymbol{\theta}_M$ for the main phase
	\caption{Warm-up phase for optimizing LCS-MPC Policy}
	\label{alg.warm_up_phase}
\end{algorithm2e}

\subsection{Model Predictive Controller with LCS}
Utilizing LCS to represent the dynamics model, one can construct a model predictive controller (MPC) as follows:
\begin{equation}\label{eqn.mpc_lcs}
\resizebox{0.91\columnwidth}{!}{$\displaystyle
\begin{aligned}
\min _{\boldsymbol{u}_t, \boldsymbol{u}_{t+1}, \dots, \boldsymbol{u}_{t + H -1}} & \sum_{k=t}^{t +H-1} \mathcal{C}\left(\boldsymbol{x}_k, \boldsymbol{u}_k\right) + \mathcal{C}_f(\boldsymbol{x}_{t+H}) \\
\text { s.t. } \quad \quad & \boldsymbol{x}_{k+1}=A \boldsymbol{x}_k+B \boldsymbol{u}_k+C \boldsymbol{\lambda}_k+\boldsymbol{d}, \\
& \mathbf{0} \leq \boldsymbol{\lambda}_k \perp D \boldsymbol{x}_k+E \boldsymbol{u}_k+F \boldsymbol{\lambda}_k+\boldsymbol{c} \geq \mathbf{0}, \\
& \boldsymbol{u}_\text{min} \leq \boldsymbol{u}_{k} \leq \boldsymbol{u}_\text{max}.
\end{aligned}$}
\end{equation}
where $H$ is the planning horizon; $\mathcal{C}$ and $\mathcal{C}_f$ are the path and terminal cost functions. And, $\boldsymbol{u}_{\text{min}}$ and $\boldsymbol{u}_{\text{max}}$ are the lower and upper bounds of actions.

Given any initial state $\boldsymbol{x}_t$, we solve the LCS-MPC in (\ref{eqn.mpc_lcs}) to plan a sequence of optimal actions $\left[\boldsymbol{u}_t, \boldsymbol{u}_{t+1}, \dots, \boldsymbol{u}_{t + H - 1}\right]$ that minimizes the total cost, then select the first action $\boldsymbol{u}_t$ to apply on the robot and repeat the process in every time step in a receding horizon manner.
To efficiently solve the LCS-MPC, we employ the direct trajectory optimization method \cite{Posa.etal2014}, which simultaneously searches over trajectories of $\boldsymbol{x}_{t:t+H}$, $\boldsymbol{u}_{t:t+H-1}$, and $\boldsymbol{\lambda}_{t:t+H-1}$, treating the LCS dynamics as a separate constraint for each time step.
Also, we use the IPOPT solver \cite{Wachter.Biegler2006} to solve such nonlinear problems.
\begin{algorithm2e}[h]
	\SetAlgoLined
    \nonl \textbf{Parameterization:} LCS-MPC policy parameters $\boldsymbol{\theta}$ and value function $\mathcal{V}_{\phi}$;\\
    \nonl \textbf{Hyper-parameters:} Total number of iterations $K$, \\
    \nonl the number of policy improvement steps $\bar{N_p}$; \\
    \nonl Learning rate $\bar{\eta}$ for the policy optimization; Loss \\
    \nonl weighting parameter $\beta$ in (\ref{eqn.total_loss}); Discount factor $\gamma$ \\
    \nonl \Add{and parameter $\zeta$} for computing the advantage values;\\
	\nonl \SetKwInput{given}{Initialization}
	\given{$\boldsymbol{\theta}_M$ obtained from the warm-up phase, $\phi_0$, and empty data buffer $\mathcal{D}$;}
	\nl\For{$k=0, 1,..., K$}
	{
		{Empty buffer $\mathcal{D}$, collect $\bar{N}$ new trajectories by running the LCS-MPC policy $\pi_{\boldsymbol{\theta}_{k}}$, and add to $\mathcal{D}$ \\
            For each trajectory, \Add{compute the generalized advantage value $\mathcal{A}_t$ as in (\ref{eqn.adv_func}), then the bootstrapped total reward $R_t = \mathcal{A}_t + \mathcal{V}_{\phi_{k}}(\boldsymbol{x}_{t})$}\\
		}
		\For{$i\gets0$ \KwTo $\bar{N_p}$}
		{
			Compute the combined loss gradient $\frac{d \mathcal{L}^{\boldsymbol{\theta}_i}_c}{d \boldsymbol{\theta}_i}$ \\
			Update $\boldsymbol{\theta}_{i+1}\leftarrow \boldsymbol{\theta}_{i} - \bar{\eta} \frac{d \mathcal{L}^{\boldsymbol{\theta}_i}_c}{d \boldsymbol{\theta}_i}$
		}
		{Fit value function $\mathcal{V}_{\phi}$ by performing regression with mean-square error
		$\displaystyle \phi_{k+1}=\arg \min _\phi \frac{1}{\left|\mathcal{D} \right| T} \sum_{\tau \in \mathcal{D}} \sum_{t=0}^T\left(\mathcal{V}_\phi\left(\boldsymbol{x}_t\right)- R_t\right)^2$ 
		}
	}
	\caption{Main phase for optimizing LCS-MPC Policy using the PPO algorithm}
	\label{alg.main_phase}
\end{algorithm2e}
\subsection{Proximal Policy Optimization}
Proximal Policy Optimization (PPO) \cite{Schulman.etal2017} is a policy gradient algorithm that focuses on determining how to make the most significant policy improvement using current data, all while avoiding excessive steps that could lead to performance collapse.
Particularly, the PPO loss is defined as follows:
\begin{equation}
	\resizebox{0.9\columnwidth}{!}{$\displaystyle
\mathcal{L}^{\boldsymbol{\theta}}_{\text{PPO}} = - \frac{1}{|\mathcal{D}| T} \sum_{\tau \in D} \sum_{t=0}^T \begin{cases}\max \left(h_t^\theta, 1-\epsilon\right) \mathcal{A}_t & \text { if } \mathcal{A}_t < 0 \\ \min \left(h_t^\theta, 1+\epsilon\right) \mathcal{A}_t & \text { if } \mathcal{A}_t \geq 0,\end{cases}$}
\label{eqn.explicit_ppo_loss}
\end{equation}
where $\mathcal{D}$ and $|\mathcal{D}|$ are the data buffer and its size, that buffer consists of on-policy rollout trajectories $\tau$, and $T$ is the length of trajectories.
There are two key quantities in (\ref{eqn.explicit_ppo_loss}): the ratio $h^{\boldsymbol{\theta}}_t = \frac{\pi_{\boldsymbol{\theta}}\left(\boldsymbol{u}_t \mid \boldsymbol{x}_t\right)}{\pi_{\boldsymbol{\theta}_{\text {old }}}\left(\boldsymbol{u}_t \mid \boldsymbol{x}_t\right)}$ and \Add{the truncated version of generalized} advantage function $\mathcal{A}_t$ \cite{Schulman.etal2016}.
Here, the ratio $h^{\boldsymbol{\theta}}_t$ indicates how much the new policy differs from the old one. The scalar $\epsilon$ defines the bounds of $h^{\boldsymbol{\theta}}_t$, which are often referred to as the trust region of policy improvements.
In addition, $\mathcal{A}_t$ guides the policy search by measuring whether a certain action is a good or bad decision within a given state. \Add{The detailed expression of $\mathcal{A}_t$ is given below}
\Add{\begin{equation}
	\label{eqn.adv_func}
	\resizebox{0.9\columnwidth}{!}{$
	\begin{aligned}
		&\mathcal{A}_t = \delta_t + (\gamma \zeta)\delta_{t+1} + (\gamma \zeta)^2\delta_{t+2} \hdots + (\gamma \zeta)^{T-t+1} \delta_{T-1}, \\
		&\;\text{with } \delta_t = r_t + \gamma \mathcal{V}_\phi\left(\boldsymbol{x}_{t+1}\right) - \mathcal{V}_\phi(\boldsymbol{x}_{t}),
	\end{aligned}$}
\end{equation}}
where $\gamma$ is the discount factor \Add{and $\zeta$ is a hyper-parameter that controls the bias-variance tradeoff of the estimation}.
The reward $r_{t}$ is obtained by executing action $\boldsymbol{u}_{t}$ at state $\boldsymbol{x}_{t}$.
The \Add{learned} value function $\mathcal{V}_{\phi}(\boldsymbol{x}_{t})$ and $\mathcal{V}_{\phi}(\boldsymbol{x}_{t+1})$ estimate the expected total rewards if we follow the current policy from state $\boldsymbol{x}_{t}$ and $\boldsymbol{x}_{t+1}$ till the end of trajectories.
\begin{figure}[t]
	\centering
	\begin{subfigure}[b]{.20\textwidth}
		\includegraphics[width=\textwidth]{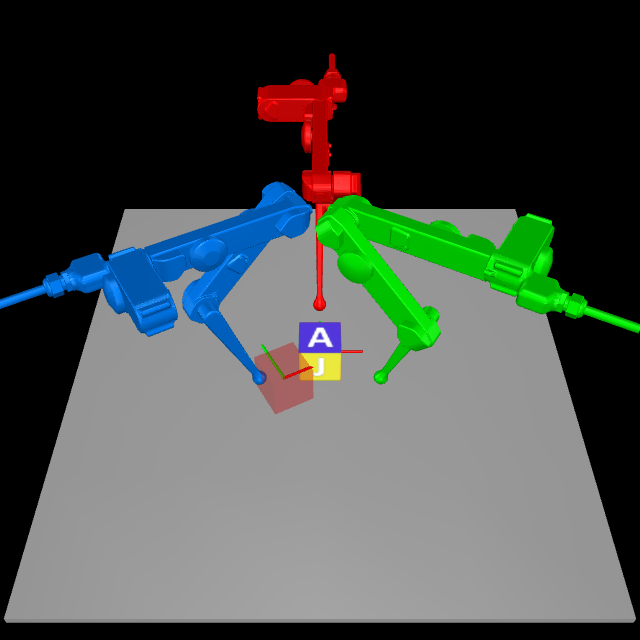} 
		\caption{}
		\label{img.trifinger}
	\end{subfigure}
        \hspace{5pt}
	\begin{subfigure}[b]{.2\textwidth}
		\captionsetup[subfigure]{labelformat=empty, font=footnotesize}
		\centering
		\begin{subfigure}{.31\textwidth}
			\includegraphics[width=\textwidth, angle=180]{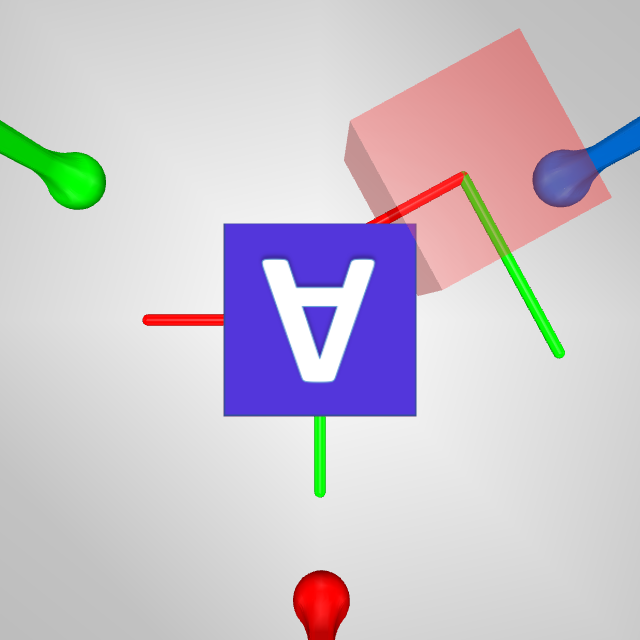}
			\caption*{Step 1}
			\label{img.traj_step_1}
		\end{subfigure}
		\hfill
		\begin{subfigure}{.31\textwidth}
			\includegraphics[width=\textwidth, angle=180]{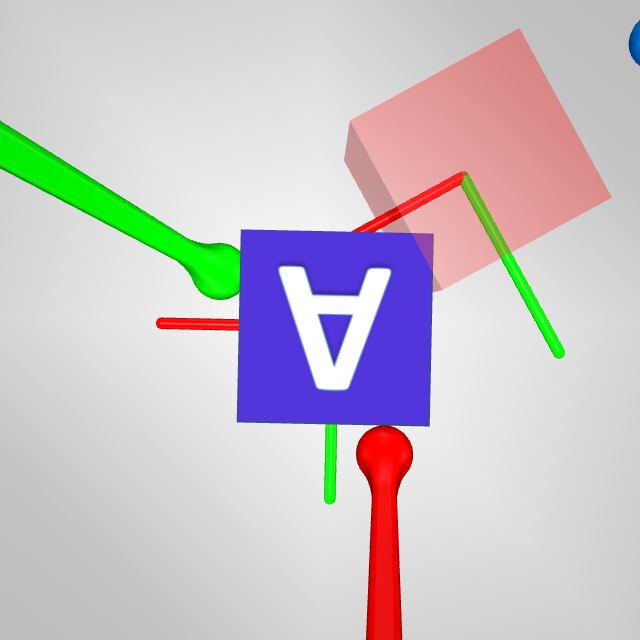}
			\caption*{Step 5}
			\label{img.traj_step_4}	
		\end{subfigure}
		\hfill
		\begin{subfigure}{.31\textwidth}
			\includegraphics[width=\textwidth, angle=180]{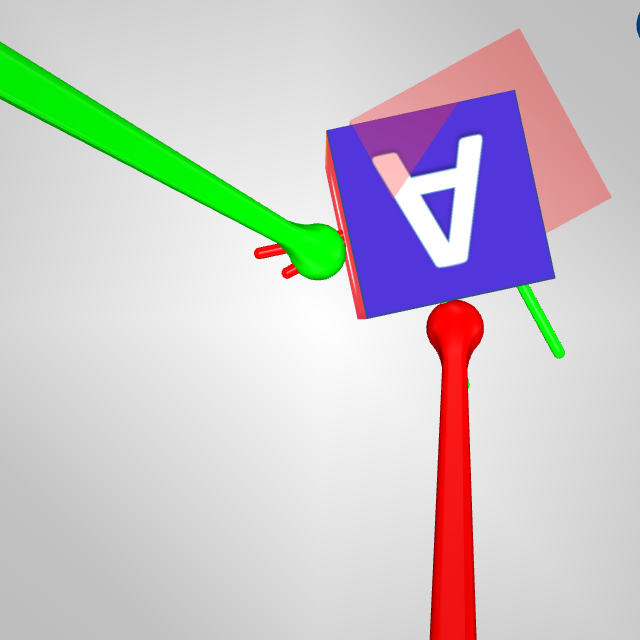}
			\caption*{Step 8}
			\label{img.traj_step_7}
		\end{subfigure}
		
		\begin{subfigure}{.31\textwidth}
			\includegraphics[width=\textwidth, angle=180]{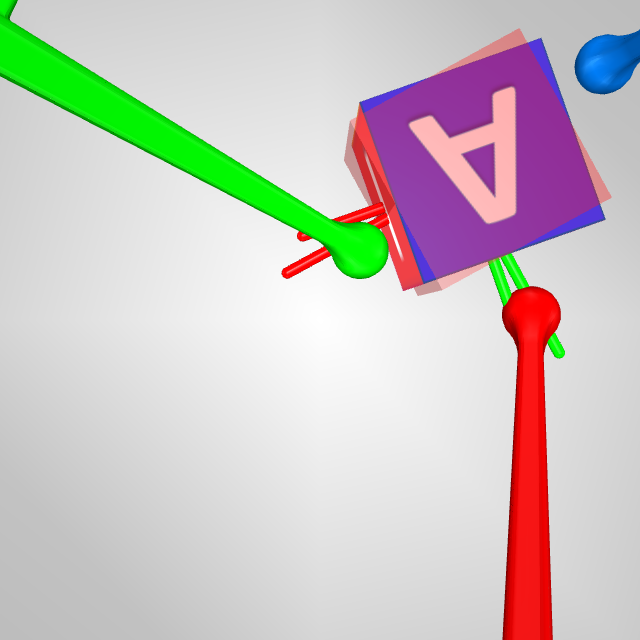}
			\caption*{Step 13}
			\label{img.traj_step_10}
		\end{subfigure}
		\hfill
		\begin{subfigure}{.31\textwidth}
			\includegraphics[width=\textwidth, angle=180]{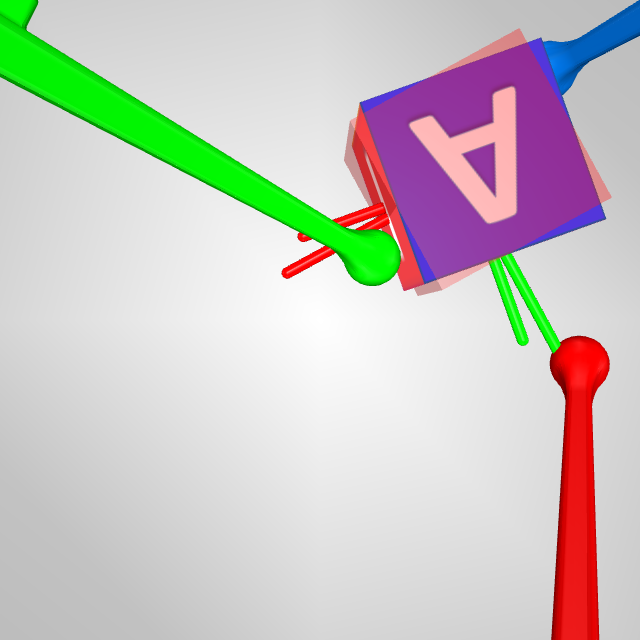}
			\caption*{Step 17}
			\label{img.traj_step_17}
		\end{subfigure}
		\hfill
		\begin{subfigure}{.31\textwidth}
			\includegraphics[width=\textwidth, angle=180]{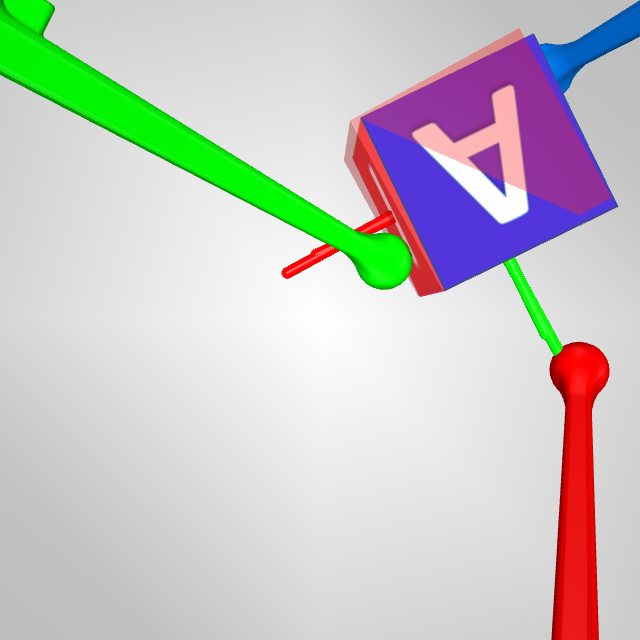}
			\caption*{Step 20}
			\label{img.traj_step_20}
		\end{subfigure}
		\caption{}
	\end{subfigure}
	
	\caption{TriFinger dexterous manipulation tasks. (a) shows the simulation environment that is constructed using MuJoCo physics engine \cite{todorov2012mujoco}. In this task, the three fingers need to push the cube towards a random target pose, visualized by the red transparent cube. (b) is an example of a rollout trajectory that demonstrates how the fingers approach, make, and break contacts to reposition the cube.}
        \vspace{-15pt}
	\label{img.trifinger_task}
\end{figure}
\section{Practical Algorithm}
In this section, we introduce our framework, LCS-RL, that utilizes a reinforcement learning algorithm, here we use PPO, to optimize an LCS dynamic model (in combination with model predictive control) for solving contact-rich tasks.

First, we formulate a stochastic policy, called the LCS-MPC stochastic policy, by adding Gaussian noise to the output of the LCS-MPC planner in (\ref{eqn.mpc_lcs}). In other words, the LCS-MPC policy is directly parameterized by the LCS model.
Then, we use the combination of the PPO loss and the violation-based loss given in (\ref{eqn.total_loss}) to improve the task performance of that policy.

We address the poor data efficiency of the PPO algorithm by leveraging data-efficient model learning at the start and then transitioning to PPO when in a good neighborhood.
Therefore, our framework consists of two phases: the warm-up phase and the main phase.
In the warm-up phase, we follow the algorithm proposed in \cite{Jin2024}, solely employing the violation-based loss (\ref{eqn.vio_loss}) to quickly learn the parameters of the LCS model that can achieve good task performance.
Subsequently, we use the learned LCS model to accelerate the main phase, where we start using the PPO algorithm.
In practice, we find that having the warm-up phase leads to more stable and progressive training than involving PPO right from the beginning.
While theoretically, we could merge the two phases and switch the loss upon transition, it is simpler to keep them separated due to different hyper-parameters required for each phase.
Details of the warm-up phase and main phase are provided in Algorithm \ref{alg.warm_up_phase} and \ref{alg.main_phase}.

\subsection{LCS-MPC Stochastic Policy}
The LCS-MPC policy is the probability density of an action distribution associated with the current state $\boldsymbol{x}_t$:
\begin{equation}
	\resizebox{0.91\columnwidth}{!}
	{$\displaystyle
\begin{aligned}
\pi_{\boldsymbol{\theta}}(\boldsymbol{u}_t | \boldsymbol{x}_t)
&= \frac{\text{exp} \left(-\frac{1}{2} (\boldsymbol{u}_t - \mu_{\boldsymbol{\Theta}}(\boldsymbol{x}_t))^T \Sigma^{-1} (\boldsymbol{u}_t - \mu_{\boldsymbol{\Theta}}(\boldsymbol{x}_t))\right)}{(2 \pi)^{n_u/2} \det(\Sigma)^{1/2}},
\end{aligned}$}
\label{eqn.stochastic_policy}
\end{equation}
where $\mu_{\boldsymbol{\Theta}} \in \mathbb{R}^{n_u}$ is \Cross{the deterministic action or} the optimal output of the LCS-MPC planner, and $\Sigma \in \mathbb{R}^{n_u \times n_u}$ is the covariance matrix that indicates the noise magnitude.
Here, $\boldsymbol{\Theta} = (A, B, C, \boldsymbol{d}, D, E, F, \boldsymbol{c})$ is actually the LCS parameters. And, $\boldsymbol{\theta} = [\boldsymbol{\Theta}, \Sigma]$ denotes the joint vector of the policy's learnable parameters.
The added noise encourages exploration, helping to avoid low-quality local minima. Typically, the noise magnitude is large initially, gradually decreasing as the policy exploits acquired knowledge for better task performance.
\begin{figure}[t]
	\centering
	\includegraphics[width=0.9\linewidth]{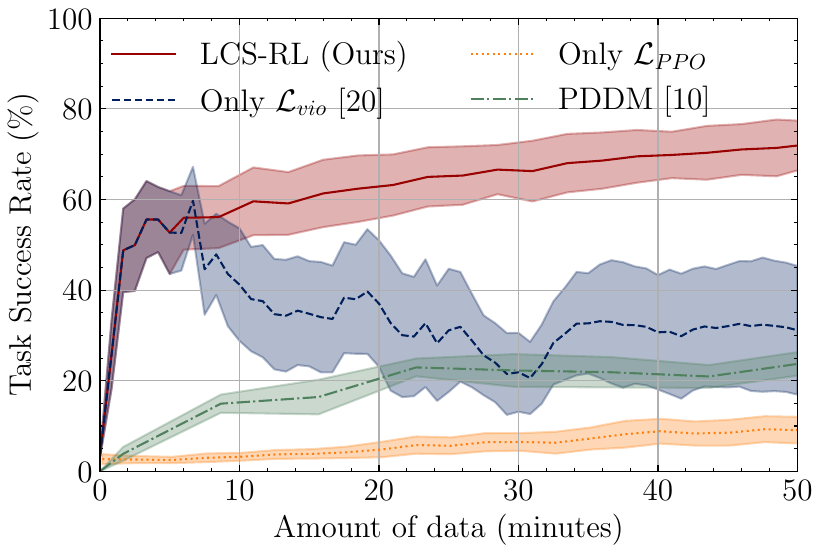}
	\caption{Learning curves of the TriFinger Moving Cube task. The red, blue, orange, and green lines show the average task success rate of our proposed method, the prior method \cite{Jin2024}, a method that uses PPO without a warm-up phase, and PDDM \cite{Nagabandi.etal2020} respectively. At the beginning of the training, our method and the prior method \cite{Jin2024} share the same performance since the same algorithm is used. However, the transition occurs after collecting 6 minutes of data, when our method switches to fully employ the PPO algorithm. Shaded regions indicate normal t-score 95\% confidence intervals.}
	\label{img.ppo_vs_vio_cube}
	\vspace{-15pt}
\end{figure}
\subsection{Loss for Optimizing LCS Model}
In our framework, we employ two types of loss: violation-based and PPO. The violation-based loss improves the forward prediction capability of the LCS model, while the PPO loss directly enhances the task performance of the LCS-MPC planner. Introducing a hyper-parameter $\beta \in [0, 1]$ allows us to balance the contributions of these losses, resulting in the combined loss:
\begin{equation}
\mathcal{L}^{\boldsymbol{\theta}}_{c} = \beta \mathcal{L}^{\boldsymbol{\theta}}_{\mathrm{PPO}} + (1 - \beta) \mathcal{L}^{\boldsymbol{\theta}}_{\mathrm{vio}},
\label{eqn.total_loss}
\end{equation}
In order to optimize the parameters of LCS-MPC stochastic policy, one must compute the gradient of the PPO loss and the violation-based loss with respect to the policy parameters. In section \ref{sec.learning_lcs} we have already mentioned the method for calculating $\frac{d \mathcal{L}^{\boldsymbol{\theta}}_{\mathrm{vio}}}{d \boldsymbol{\theta}}$.
Moving forward, we will illustrate the process for computing $\frac{d \mathcal{L}^{\boldsymbol{\theta}}_{\mathrm{PPO}}}{d \boldsymbol{\theta}}$.

\subsection{Gradient of PPO Loss}
The original PPO algorithm parameterizes policies via deep neural networks \cite{Schulman.etal2017}, thus computing the gradient of the PPO loss over policy parameters can be simply done via automatic differentiation.
However, it does not apply to our case since our policy is actually an MPC planner. Differentiating through the MPC requires special treatment.

Given the explicit form of PPO loss in (\ref{eqn.explicit_ppo_loss}), using chain rule, the gradient with respect to $\boldsymbol{\theta}$ can be computed
\begin{equation}
\label{eqn.dloss_dtheta}
\resizebox{\columnwidth}{!}
{
$\displaystyle \frac{d \mathcal{L}^{\boldsymbol{\theta}}_{\mathrm{PPO}}}{d \boldsymbol{\theta}} = \frac{-1}{|\mathcal{D}| T} \sum_{\tau \in D} \sum_{t=0}^T \begin{cases}\frac{d h_t^{\boldsymbol{\theta}}}{d \pi_{\boldsymbol{\theta}}} \frac{d \pi_{\boldsymbol{\theta}}}{d \boldsymbol{\theta}} \mathcal{A}_t & \hspace{-5pt} \text {if } h_t^{\boldsymbol{\theta}} \geq 1-\epsilon \text {; } \mathcal{A}_t<0 \\ 
0 & \hspace{-5pt} \text {if } h_t^{\boldsymbol{\theta}} < 1-\epsilon \text {; } \mathcal{A}_t<0 \\ 
\frac{d h_t^{\boldsymbol{\theta}}}{d \pi_{\boldsymbol{\theta}}} \frac{d \pi_{\boldsymbol{\theta}}}{d \boldsymbol{\theta}} \mathcal{A}_t & \hspace{-5pt} \text {if } h_t^{\boldsymbol{\theta}} \leq 1 + \epsilon \text {; } \mathcal{A}_t \geq 0 \\
0 & \hspace{-5pt} \text {if } h_t^{\boldsymbol{\theta}} > 1 + \epsilon \text {; } \mathcal{A}_t \geq 0.\end{cases}$
}
\end{equation}
Due to the clipping effect of the PPO loss, the gradients are zero when the improvement steps of the PPO policy $h^{\boldsymbol{\theta}}_t$ are outside of trust region $[1 - \epsilon, 1 + \epsilon]$. Hence, we are left to compute the gradients if $h^{\boldsymbol{\theta}}_t$ stays within the trust region. To compute (\ref{eqn.dloss_dtheta}), one must compute $\frac{d h_t^{\boldsymbol{\theta}}}{d \pi_{\boldsymbol{\theta}}}$ and $\frac{d \pi_{\boldsymbol{\theta}}}{d \boldsymbol{\theta}}$.
While it is straightforward to evaluate $\frac{d h_t^{\boldsymbol{\theta}}}{d \pi_{\boldsymbol{\theta}}}$, computing $\frac{d \pi_{\boldsymbol{\theta}}}{d \boldsymbol{\theta}}$ requires differentiation of the optimal actions of the MPC $\mu_{\boldsymbol{\Theta}}(\boldsymbol{x}_t)$ with respect to its parameters $\boldsymbol{\Theta}$.
We compute this derivative via perturbations of KKT conditions \cite{kuhn1951nonlinear} with details given in \cite{Bskens1998SensitivityAA}.
Also, note that the MPC problem is not always classically differentiable (e.g. when strict complementarity does not hold in the KKT conditions), but we have not found this problematic in practice.
 \begin{table}[t!]
	\centering
	\resizebox{\columnwidth}{!}{
		\begin{tabular}{|c|c|c|c|c|}
			\hline
			\multirow{2}{*}{} & \multicolumn{2}{c|}{\textbf{Peak Success Rate (\%)}} & \multicolumn{2}{c|}{\textbf{Final Success Rate (\%)}} \\
			\cline{2-5}
			\textbf{Object} & \multirow{2}{*}{\textbf{Only} $\mathbf{\mathcal{L}_{\mathrm{vio}}^{\theta}}$} & \textbf{LCS-RL} & \multirow{2}{*}{\textbf{Only} $\mathbf{\mathcal{L}_{\mathrm{vio}}^{\theta}}$} & \textbf{LCS-RL} \\
			& & \textbf{(Ours)} & & \textbf{(Ours)} \\ \hline
			Sugar Box
			& $87.5 \pm 9.2$ 
			& $\boldsymbol{95.9 \pm 2.4}$
			& $44.3 \pm 26.2$ 
			& $\boldsymbol{95.9 \pm 2.4}$ \\ \hline
			Fish Can 
			& $59.1 \pm 7.3$ 
			& $\boldsymbol{69.9 \pm 8.1}$ 
			& $38.8 \pm 13.1$ 
			& $\boldsymbol{69.9 \pm 8.1}$ \\ \hline
			Mug
			& $32.5 \pm 8.4$ 
			& $\boldsymbol{44.6 \pm 3.9}$ 
			& $10.0 \pm 6.3$ 
			& $\boldsymbol{44.6 \pm 3.9}$ \\ \hline
			Wrench
			& $45.5 \pm 8.7$ 
			& $\boldsymbol{60.0 \pm 6.5}$
			& $11.5 \pm 8.6$ 
			& $\boldsymbol{59.5 \pm 6.6}$ \\ \hline
			Clamp
			& $24.7 \pm 5.3$ 
			& $\boldsymbol{39.3 \pm 9.7}$ 
			& $6.1 \pm 5.6$ 
			& $\boldsymbol{38.7 \pm 10.2}$ \\ \hline
			Banana
			& $28.4 \pm 5.8$ 
			& $\boldsymbol{35.5 \pm 5.3}$ 
			& $7.5 \pm 6.6$ 
			& $\boldsymbol{35.5 \pm 5.3}$ \\
			\hline
		\end{tabular}
	}
	\caption{Comparison task success rates between our method and prior method \cite{Jin2024} on diverse objects.}
	\vspace{-10pt}
	\label{table.objects}
\end{table}
\section{Experiments and Results}
In this section, we will verify our proposed framework on the three-fingered robotic hand manipulation task that was first proposed by \cite{Jin2024} (see Fig. \ref{img.trifinger_task}). We call it the TriFinger Moving Cube task.
In the first experiment, we show a comparison of the task performance of the LCS model trained by our method and prior methods.
Next, we replace the cube with other objects that have more complex shapes and repeat the same experiment.
Lastly, we demonstrate that the data efficiency of our framework can be greatly improved via transfer learning.
In order to guarantee statistically meaningful results, for each experiment, we have 10 runs with 10 random seeds. Also, we compute the task success rate by evaluating the learned models with 1000 random goal poses and aggregate results.
All videos and codes are available at \href{https://sites.google.com/view/lcs-rl}{https://sites.google.com/view/lcs-rl}.

\subsection{TriFinger Moving Cube Task}

In this task, a TriFinger robot aims to align a 6 cm-sized cube with random goal poses on a planar surface. Each episode comprises up to 20 steps, each lasting 0.1 seconds.
\begin{figure*}[t!]
	\centering
	\begin{subfigure}[c]{.37\textwidth}
		\centering
		\begin{subfigure}{0.33\textwidth}
			\raisebox{-26pt}{\includegraphics[width=\textwidth]{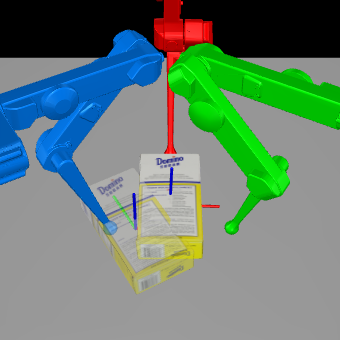}}
		\end{subfigure}
		\hfill
		\begin{subfigure}[c]{.65\textwidth}
			\includegraphics[width=\textwidth]{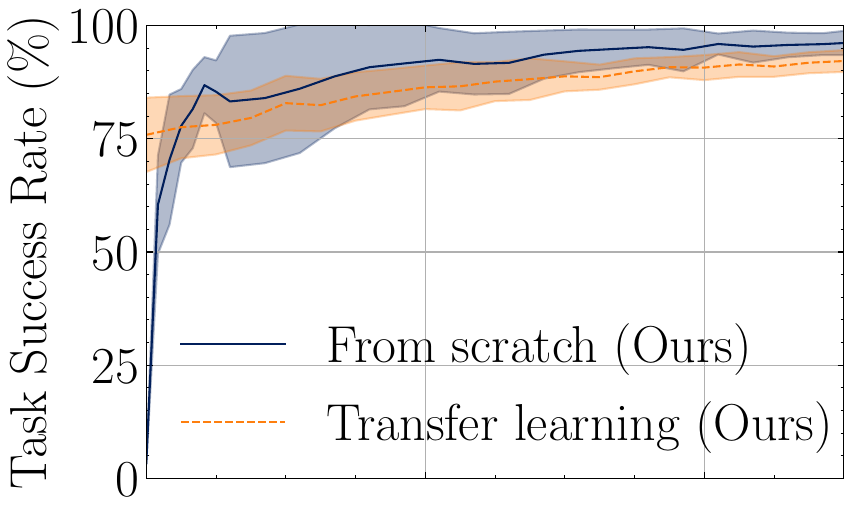}
		\end{subfigure}
            \caption{Sugar Box}
	\end{subfigure}
        \hspace{10pt}
 	\begin{subfigure}[c]{.37\textwidth}
		\centering
		\begin{subfigure}{0.33\textwidth}
			\raisebox{-26pt}{\includegraphics[width=\textwidth]{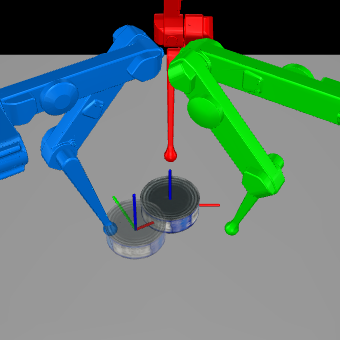}}
		\end{subfigure}
		\hfill
		\begin{subfigure}[c]{.65\textwidth}
			\includegraphics[width=\textwidth]{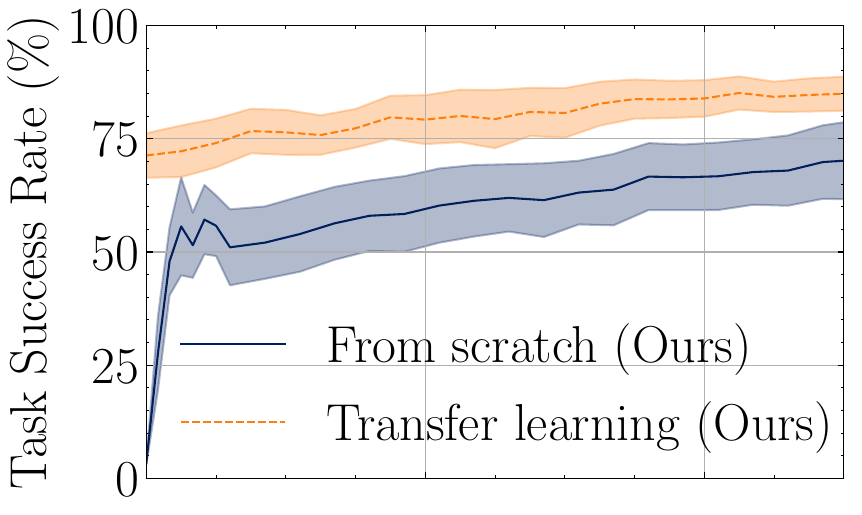}
		\end{subfigure}
            \caption{Fish Can}
	\end{subfigure}
 	\begin{subfigure}[c]{.37\textwidth}
		\centering
		\begin{subfigure}{0.33\textwidth}
			\raisebox{-26pt}{\includegraphics[width=\textwidth]{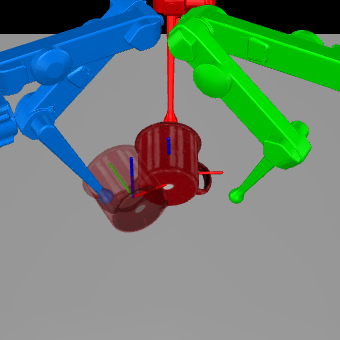}}
		\end{subfigure}
		\hfill
		\begin{subfigure}[c]{.65\textwidth}
			\includegraphics[width=\textwidth]{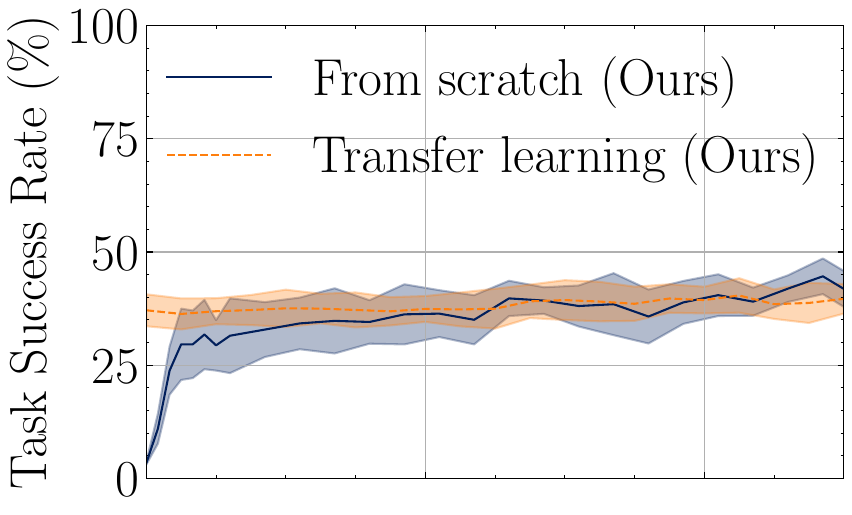}
		\end{subfigure}
            \caption{Mug} 
	\end{subfigure}
        \hspace{10pt}
  	\begin{subfigure}[c]{.37\textwidth}
		\centering
		\begin{subfigure}{0.33\textwidth}
			\raisebox{-26pt}{\includegraphics[width=\textwidth]{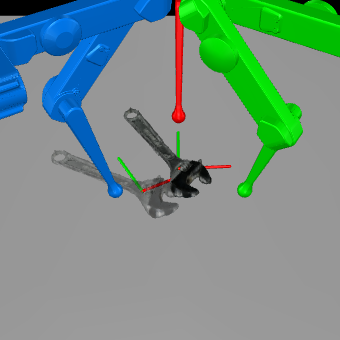}}
		\end{subfigure}
		\hfill
		\begin{subfigure}[c]{.65\textwidth}
			\includegraphics[width=\textwidth]{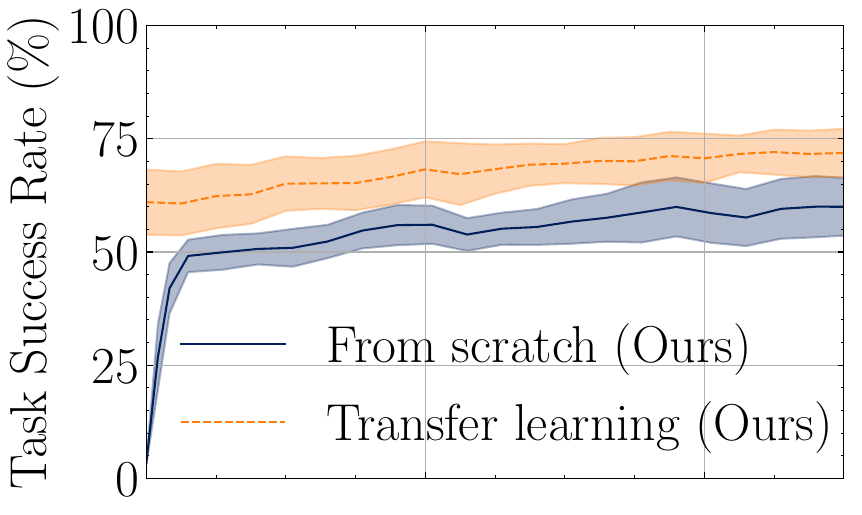}
		\end{subfigure}
            \caption{Wrench}
	\end{subfigure}
         \begin{subfigure}[c]{.37\textwidth}
		\centering
		\begin{subfigure}{0.33\textwidth}
			\raisebox{-21pt}{\includegraphics[width=\textwidth]{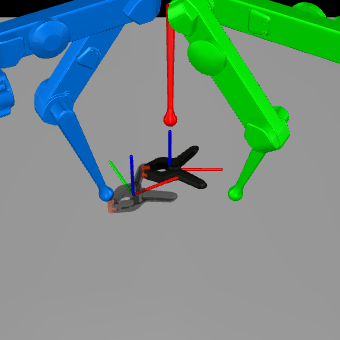}}
		\end{subfigure}
		\hfill
		\begin{subfigure}[c]{.65\textwidth}
			\includegraphics[width=\textwidth]{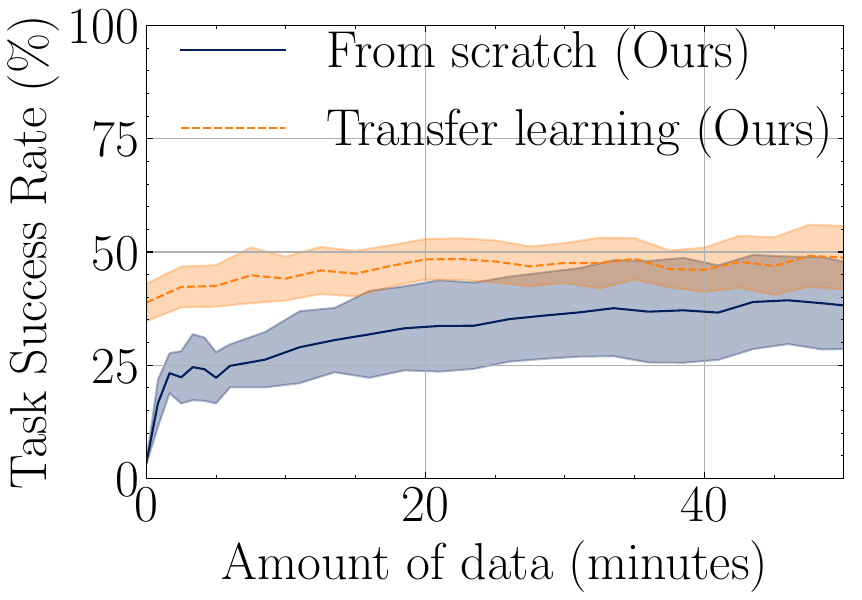}
		\end{subfigure}
            \caption{Clamp}
	\end{subfigure}
        \hspace{10pt}
            \begin{subfigure}[c]{.37\textwidth}
		\centering
		\begin{subfigure}{0.33\textwidth}
			\raisebox{-21pt}{\includegraphics[width=\textwidth]{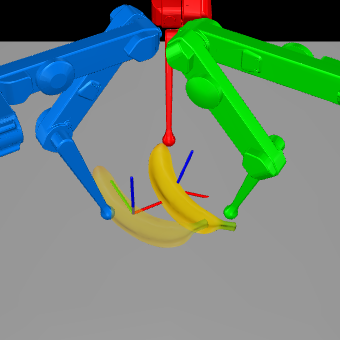}}
		\end{subfigure}
		\hfill
		\begin{subfigure}[c]{.65\textwidth}
			\includegraphics[width=\textwidth]{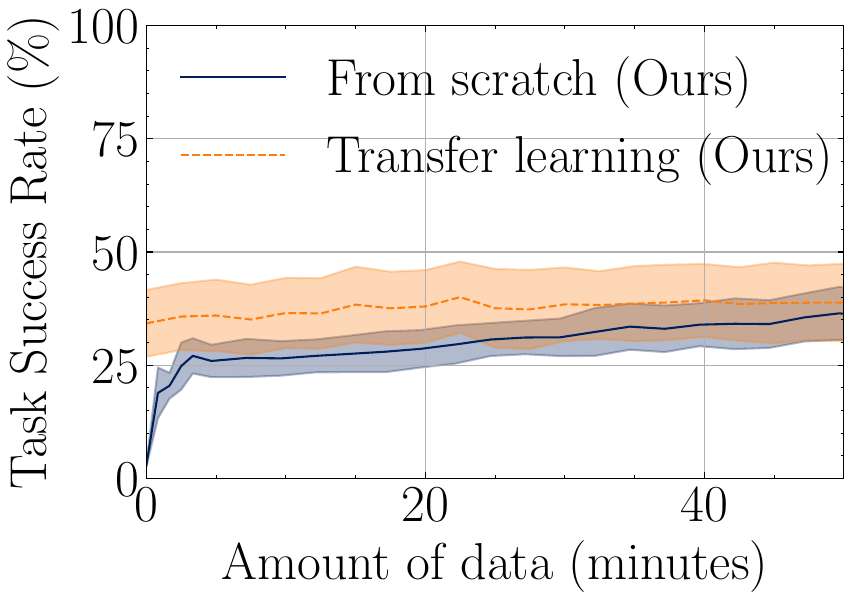}
		\end{subfigure}
            \caption{Banana}
	\end{subfigure}
        \caption{Comparison task performance between learning the LCS model from scratch and pre-trained models (obtained from training with the TriFinger Moving Cube task) on the YCB objects using our LCS-RL framework.}
        \label{fig.transfer_learning}
        \vspace{-15pt}
\end{figure*}
\subsubsection{States and Actions}

We define the system state
\begin{equation}
	\boldsymbol{x} = [\boldsymbol{p}_{\text{cube}},\;\alpha_{\text{cube}},\;\boldsymbol{p}_{\text{fingertips}}] \in \mathbb{R}^{9},
\end{equation}
where $\boldsymbol{p}_{\text{cube}} \in \mathbb{R}^{2}$ is the xy position of the cube; $\alpha_{\text{cube}} \in \mathbb{R}$ is the rotation angle around the z (vertical) axis; and $\boldsymbol{p}_{\text{fingertips}} \in \mathbb{R}^6$ are the xy positions of three fingertips. We define the actions as deviations from the current positions of three fingertips in the Cartesian space. In addition, we impose safety limits on actions (element-wise) to constrain how far fingertips can move in one time step
\begin{equation}
    \begin{aligned}
        &\boldsymbol{u} = \Delta \boldsymbol{p}_{\text{fingertips}} \in \mathbb{R}^6, \\
        & u_i \in [-0.015,\; 0.015] \;\; \text{m}.
    \end{aligned}
\end{equation}
We employ operational space control (OSC) \cite{Khatib1987AUA} in the lower-level controller to map action $\boldsymbol{u}$ to the joint torque of each finger. We also utilize the OSC controller to maintain fingertips at a constant height as this TriFinger task involves only planar manipulation.

\subsubsection{Task Space}
We use the same bounds for task space as in \cite{Jin2024}, from which the goal poses are uniformly sampled
\begin{equation}
    \begin{aligned}
        [-0.06,\; -0.06]^T\; \text{m} &\leq \boldsymbol{p}_{\text{goal}} \leq [0.06,\; 0.06]^T\; \text{m}, \\
        -0.5\; \text{rads} &\leq \alpha_{\text{goal}} \leq 0.5\; \text{rads}.
    \end{aligned}
\end{equation}

\subsubsection{Task Success Criteria}
When the cube pose is near the goal pose and within some tolerances, we can consider that the task is successfully completed.
The goal tolerance values are selected to establish the right level of difficulty as too stringent tolerances make the task impossible to solve.
We follow some previous works on TriFinger tasks \cite{Allshire.etal2021, Funk.etal2022} to set the tolerances as follows:
\begin{equation}
	\begin{aligned}
		&\|\boldsymbol{p}_{\text{cube}}^{\ast} - \boldsymbol{p}_{\text{goal}}\| \leq 0.02\; \text{m}, \\
		&\|\alpha_{\text{cube}}^{\ast} - \alpha_{\text{goal}}\| \leq 0.2\; \text{rads},
	\end{aligned}
\end{equation}
where $\boldsymbol{p}_{\text{cube}}^{\ast}$ and $\alpha_{\text{cube}}^{\ast}$ are the xy position and orientation of the cube at the last time step $T$; $ \boldsymbol{p}_{\text{goal}} \in \mathbb{R}^2$ and $\alpha_{\text{goal}} \in \mathbb{R}$ together specify the goal pose.

\subsubsection{Cost function for the LCS-MPC}
We utilize the same cost function for the LCS-MPC as in the prior work \cite{Jin2024}:
\begin{equation}
\resizebox{0.89\columnwidth}{!}{$\displaystyle
\begin{aligned}
	& \qquad \qquad \quad \mathcal{J} = \sum_{k=t}^{t+H-1} \mathcal{C}(\boldsymbol{x}_k, \boldsymbol{u}_k) + \mathcal{C}_f(\boldsymbol{x}_{t+H}), \\
	& \mathcal{C} = 10.0\left\|\boldsymbol{p}_{\text {fingertips }}-\boldsymbol{p}_{\text {cube }}\right\|^2+ 200.0\left\|\boldsymbol{p}_{\text {cube }}-\boldsymbol{p}_{\text {goal }}\right\|^2 \\
	& \quad + 0.3\left(\alpha_{\text {cube }}-\alpha_{\text {goal }}\right)^2 + 200.0\left\|\boldsymbol{u}\right\|^2, \\
	& \mathcal{C}_f = 6.0\left\|\boldsymbol{p}_{\text {fingertips }}-\boldsymbol{p}_{\text {cube }}\right\|^2+200.0\left\|\boldsymbol{p}_{\text {cube }}-\boldsymbol{p}_{\text {goal }}\right\|^2 \\
	& \quad + 1.5\left(\alpha_{\text {cube }}-\alpha_{\text {goal }}\right)^2, \\
	\end{aligned}
	\label{eqn.cost_function}$}
\end{equation}

\subsubsection{Reward function for PPO}
We use both dense and sparse reward functions for the PPO algorithm. The dense reward function $r_t(\boldsymbol{x}_t, \boldsymbol{u}_t)$ is simply the negation of the cost function $\mathcal{C}$ in (\ref{eqn.cost_function}). This choice of reward function ensures that both PPO and the MPC planner of the stochastic policy align in the same direction toward task completion.

At the end of the rollout trajectory, we add a negative sparse reward to penalize for not completing the task:
\begin{equation}
\begin{aligned}
& r(\boldsymbol{x}_{T-1}, \boldsymbol{u}_{T-1}) = -10.0 \times (1 - \textit{is\_task\_completed}), \\
\end{aligned}
\end{equation}
In practice, we find that sparse reward helps to accelerate the training significantly.

\subsection{Results of the TriFinger Moving Cube Task}
To demonstrate the effectiveness of LCS-RL, we compare against the prior method \cite{Jin2024}, which trains purely on $\mathcal{L}^{\boldsymbol{\theta}}_{\mathrm{vio}}$, against PPO without a warm-up phase. We also compare against a state-of-the-art model-based RL approach PDDM \cite{Nagabandi.etal2020} to demonstrate the utility of simple, non-smooth models over deep neural networks for manipulation.
Note that in the main phase of our method, we set $\beta = 1.0$ for the combined loss in (\ref{eqn.total_loss}), meaning that only the PPO loss is used.
The results are shown in Fig.\ref{img.ppo_vs_vio_cube}.

When utilizing only the violation-based loss, the mean success rate peaks at 55\% after 7 minutes of data, then fluctuates and decreases as more data is collected, and finally stops at approximately 30\%.
In contrast, starting with the same performance at 6 minutes of data, our method improves the task performance throughout the training, reaching 65\% of success rate after 25 minutes of data and 71.4\% at the end of the training.
Since the LCS models have limited expressiveness power, even if we optimize LCS models for better capability of forward prediction, this capability might not be optimally assigned to regions of state space where accuracy is needed for task performance.
As a result, the overall task performance might drop significantly.
Our method does not suffer from that issue because the only objective of PPO is encouraging the policy, to repeat good trajectories and avoid bad trajectories.

In addition, the PPO-only method and PDDM \cite{Nagabandi.etal2020} have the lowest task performances throughout the training, achieving merely $10\%$ and $20\%$ for the final task success rate.

\subsection{TriFinger Manipulating Diverse Objects}
We run a set of experiments on the TriFinger Moving Object task, which is similar to the TriFinger Moving Cube task, but the cube is replaced by other objects with non-convex, highly intricate shapes.
Those objects, including sugar box, fish can, mug, wrench, clamp, and banana, are selected from the YCB object and model set \cite{Calli.etal2015}.
As seen from Table \ref{table.objects}, our method consistently outperforms the prior method in \cite{Jin2024} given the same amount of data, gaining from 8\% (sugar box) to 15\% (clamp) higher task success rate. 

\subsection{Transfer Learning}
To illustrate the transfer learning capabilities of our LCS-RL framework, we employ the LCS model initially trained on the TriFinger Moving Cube task as the starting point for training on other objects.
The results in Fig. \ref{fig.transfer_learning} show that our LCS-RL framework is highly suitable for transfer learning. Particularly, we can observe that transfer learning significantly accelerates the training, yielding even higher final task success rates in all objects (except for the sugar box), compared to the training from scratch model.

\section{Conclusions}
In conclusion, we present LCS-RL, a novel approach that leverages a reinforcement learning algorithm to directly maximize the task performance of the LCS model in combination with the MPC planner.
We demonstrate that the proposed method attains higher task performance and greater sample efficiency compared to prior methods in TriFinger robot tasks involving pushing and rotating various objects.
In addition, we show that our method is highly suitable for transfer learning, which further helps to improve data efficiency.

Our framework is not limited to only the PPO algorithm since any RL algorithms can be incorporated.
Thus, one direction for future work is to explore other RL algorithms and employ them in our framework.
There are off-policy RL algorithms such as TD3 \cite{Fujimoto.etal2018} or SAC \cite{Haarnoja.etal2018}, known for better data efficiency when compared to on-policy algorithms \cite{Lillicrap.etal2019}.
Nevertheless, this advantage may not be as evident in situations with limited data.

Lastly, we observe the limitation of using the LCS model for representing system dynamic models, especially with complex geometries like bananas or clamps. We aim to investigate alternative structured models simpler than neural networks yet exhibiting nonlinear components, although with a trade-off in data efficiency. One potential candidate is the Nonlinear Complementarity System (NCS) model.


\bibliographystyle{IEEEtran}
\bibliography{lcs_rl}{}

\end{document}